\documentclass[11pt]{article}

\usepackage[margin=0.9in]{geometry}
\usepackage{graphicx}
\usepackage{placeins}

\usepackage{booktabs}
\usepackage{amsmath}
\usepackage{amssymb}
\usepackage{array}
\usepackage{tabularx}
\usepackage{xcolor}
\usepackage{colortbl}
\usepackage{xurl}
\usepackage{needspace}
\usepackage{microtype}
\usepackage{titlesec}
\usepackage{fancyhdr}
\usepackage{caption}
\usepackage{subcaption}
\usepackage{enumitem}
\usepackage{acro}
\usepackage[most]{tcolorbox}
\usepackage{hyperref}
\usepackage{wrapfig}
\usepackage[capitalize,noabbrev,nameinlink]{cleveref}
\definecolor{numsblue}{HTML}{000000}
\definecolor{numsaccent}{HTML}{2F7D63}
\definecolor{numslight}{HTML}{F2F2F2}
\definecolor{numsborder}{HTML}{D0D0D0}
\definecolor{numsgrey}{HTML}{5F6B7A}

\newcolumntype{Y}{>{\raggedright\arraybackslash}X}
\newcommand{\nums}{Nums AI}
\newcommand{\methodname}{Causilo}
\newcommand{\method}{\texorpdfstring{\mbox{\methodname}}{\methodname}}

\newcommand{\mypar}[1]{\textbf{#1}\hspace{4pt}}
\DeclareAcronym{tfm}{
  short = TFM,
  long = tabular foundation model
}
\newcommand{\reportwordmark}{\includegraphics[height=10pt]{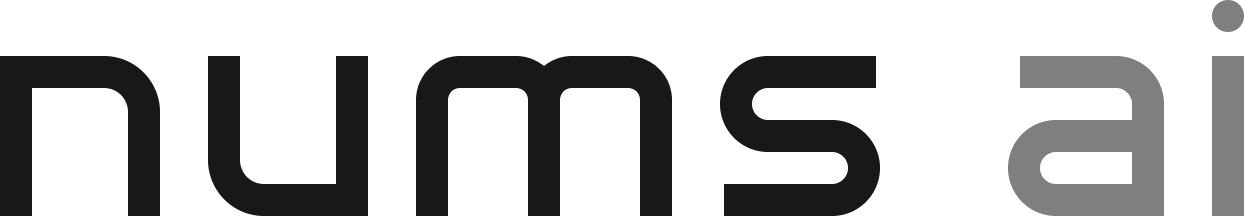}}
\newcommand{\reporttitle}{\method{} Technical Report}

\titleformat{\section}
  {\large\bfseries\color{numsblue}}
  {\thesection}{0.65em}{}
\titlespacing*{\section}{0pt}{1.25em}{0.45em}
\titleformat{\subsection}
  {\normalsize\bfseries\color{numsblue}}
  {\thesubsection}{0.65em}{}
\titlespacing*{\subsection}{0pt}{1em}{0.35em}
\setlist{nosep,leftmargin=*}
\renewcommand{\arraystretch}{1.12}
\fancypagestyle{firstpage}{%
  \fancyhf{}
  \fancyfoot[C]{\small\thepage}

}

\hypersetup{
  colorlinks=true,
  linkcolor=numsblue,
  citecolor=numsblue,
  urlcolor=numsblue,
  pdfauthor={Nums AI},
  pdftitle={\reporttitle}
}

\renewenvironment{abstract}{%
  \begin{tcolorbox}[
    enhanced,
    colback=numslight,
    colframe=numsborder,
    boxrule=0.5pt,
    arc=1mm,
    left=10pt,
    right=10pt,
    top=12pt,
    bottom=12pt
  ]
  {\raggedright
    {\Huge\bfseries
      \mbox{%
        \reporttitle
      }\par}
    \vspace{0.7em}
    {\reportwordmark\hspace{0.6em}{\small (see \hyperref[sec:contributors]{Appendix~\ref*{sec:contributors}} for the list of contributors)}\par}
  \par}
  \vspace{1.2em}
  {\color{numsborder}\hrule height 0.6pt}
  \vspace{1.2em}
  \normalsize\textbf{Abstract.}\hspace{4pt}
}{%
  \end{tcolorbox}
}

\title{\reporttitle}
\author{\nums}
\date{}

\begin{document}
\thispagestyle{firstpage}

\begin{abstract}
We introduce \textbf{\method{}}, a \ac{tfm} that combines frontier predictive performance with exceptionally fast inference. On TabArena, \method{} achieves 1785.4 Elo, at a median inference time of 0.10 seconds per 1K test samples. It outperforms TabPFN-3.5-Fast with 31.6\% less inference time, placing it on the performance--efficiency Pareto frontier. \method{} follows TabICL's column-then-row architecture but introduces another row-refinement module before row compression. This module exchanges information among cell representations within each row after column encoding. The refined cells then visit the context set again through an additional column stage before being compressed into row embeddings. For inference efficiency, both row stages use cross-attention through a fixed number of summary tokens, keeping their attention cost linear in the number of features. 
Pretrained on approximately 36M synthetic tables, \method{} delivers strong benchmark results across TabArena, BeyondArena, and ScoringBench,
achieving frontier-level performance with substantially faster inference.
\end{abstract}

\par\vspace{0.4em}
\begingroup
\renewcommand{\arraystretch}{1.3}
\setlength{\arrayrulewidth}{2pt}
\begin{tabular}{|@{\hspace{10pt}}p{0.19\textwidth}@{\hspace{12pt}}l@{}}
  \textsc{Date} & 18 Sep 2026 \\
  \textsc{Model} & 
  \href{https://huggingface.co/nums-ai/causilo}{\texttt{https://huggingface.co/nums-ai/causilo}} \\
  \textsc{Code} & \href{https://github.com/nums-ai/causilo}{\texttt{https://github.com/nums-ai/causilo}}
\end{tabular}
\par
\endgroup

\begin{figure}[!b]
  \centering
  \setlength{\parskip}{0pt}
  \includegraphics[draft=false,width=\textwidth,height=0.38\textheight,keepaspectratio]{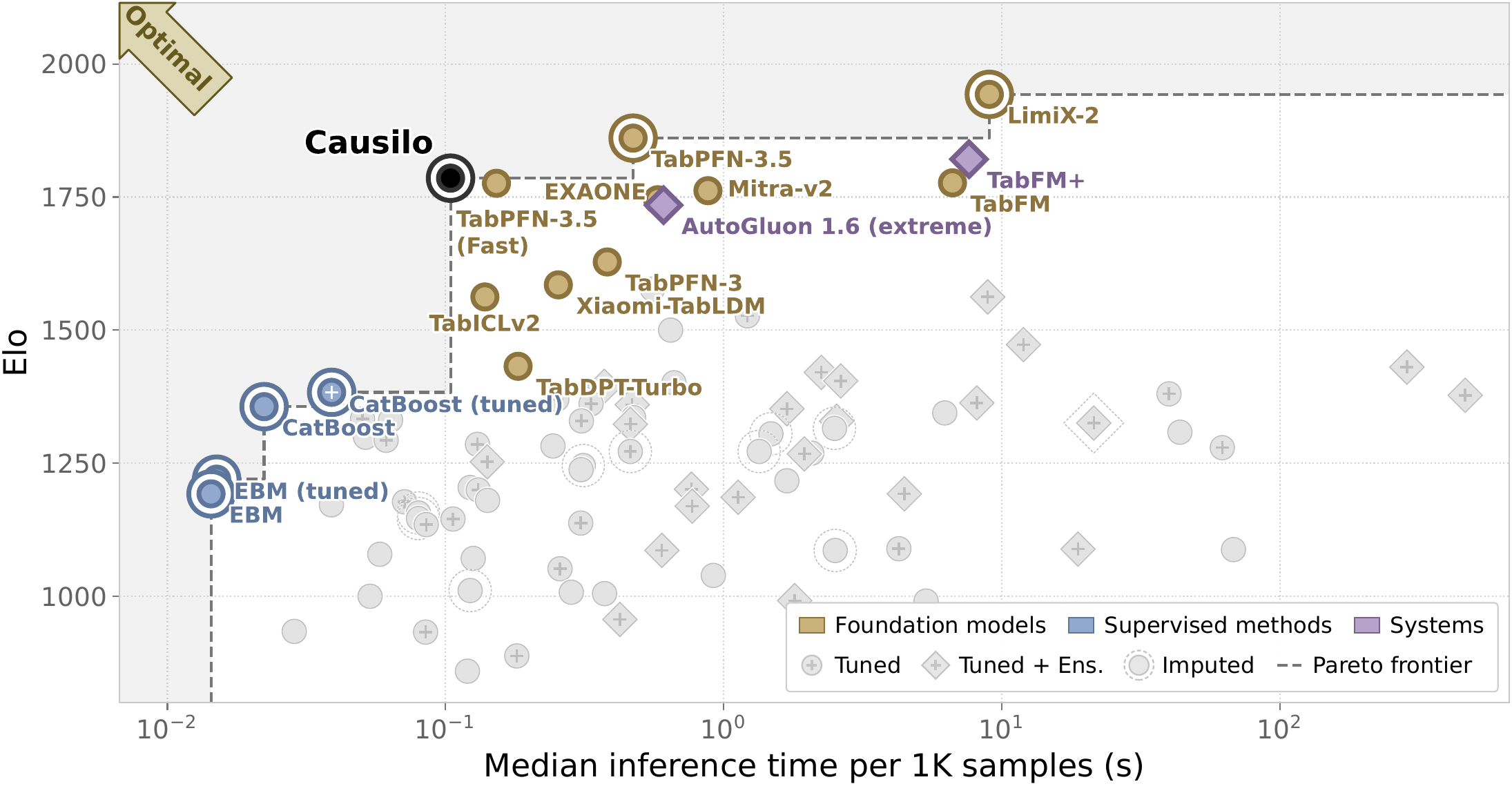}
  \caption{\textbf{TabArena performance--efficiency frontier.} The horizontal axis shows median inference time per 1K test samples on a logarithmic scale. \method{} achieves 1785 Elo at 0.1043 seconds, placing it on the Pareto frontier. Its classification and regression models contain 36.08M and 37.08M parameters, roughly one-sixth as many as TabPFN-3.5 (218.98M) and one-eleventh as many as LimiX-2 (406.2M), which also lie on the frontier.}
  \label{fig:overview}
\end{figure}

\clearpage

\section{Introduction}
\label{sec:introduction}
\acreset{tfm}

We introduce \textbf{\method{}}, a \ac{tfm} that combines strong predictive performance with outstandingly fast inference.
Its classification and regression models contain 36.08M and 37.08M parameters, respectively.
On TabArena~\cite{erickson2026tabarena}, \method{} surpasses the previous state-of-the-art TabFM~\cite{google2026tabfm} in Elo while running faster than TabICLv2~\cite{qu2026tabiclv2}, previously the fastest TFM on the Pareto frontier.

\Acp{tfm} are pretrained on diverse synthetic tasks to learn a reusable prediction procedure that transfers across datasets.
At inference, they condition on labeled \emph{context} examples to predict the targets of \emph{query} examples without updating their parameters.
TabPFN~\cite{hollmann2022tabpfn} and TabPFN-2~\cite{hollmann2025accurate} established this paradigm as a leading approach to tabular prediction.
Modern \acp{tfm} largely use Transformer architectures~\cite{vaswani2017attention}, but differ in how they organize computation across features and samples.\footnote{
We use \emph{features} and \emph{columns}, as well as \emph{samples} and \emph{rows}, interchangeably throughout the paper.
We use \emph{cell tokens} or \emph{cell representations} for the vectors processed before row compression, and \emph{row embeddings} or \emph{row representations} for the vectors produced by compression.
}

One line of work maintains cell representations throughout the network.
TabPFN-2.5~\cite{grinsztajn2025tabpfn} and LimiX~\cite{zhang2025limix}, for example, alternate attention across features and samples.
Cell tokens exchange information within each row and then incorporate information from context rows within each column.
This preserves rich representations but incurs quadratic attention costs in both the number of features and the number of samples.
Recent models such as EXAONE-Tabular~\cite{eo2026exaone} retain this design, repeatedly updating cell tokens through attention along both the feature and sample axes.

A second line of work compresses each row into a fixed-width embedding, then uses an in-context learning (ICL) Transformer to predict query targets from labeled context rows.
TabICL~\cite{qu2025tabicl} was the first to introduce this column-then-row architecture.
It first updates cell tokens across samples within each column, then compresses the resulting representations into a fixed-width embedding for each row.
A separate ICL Transformer processes these row embeddings, so its attention cost no longer depends on the number of features.
TabICLv2~\cite{qu2026tabiclv2} and TabPFN-3~\cite{grinsztajn2026tabpfn} follow the same general design.
This makes cross-row processing more efficient, but subsequent layers can no longer update individual cell tokens.

These architectural choices result in a clear performance--efficiency trade-off.
On TabArena, column-then-row models provide low inference latency, while models that retain cross-axes attention achieve stronger Elo at greater inference cost.
Dataset-specific fine-tuning, as used by Mitra-v2~\cite{tao2026mitra}, offers another route to stronger predictions, but requires additional optimization for every new dataset.
Therefore, the resulting frontier leaves an important architectural question:
\emph{can a \ac{tfm} retain richer cell representations while preserving the efficiency of in-context learning over row embeddings?}

\method{} addresses this trade-off by introducing an additional refinement module before row compression.
The model first performs target-aware column encoding to construct cell representations informed by the context set.
A row-refinement module then exchanges information among cell tokens within each row through a small set of summaries that persist across its layers.
Rather than immediately compressing the refined cell tokens into a row embedding, \method{} sends them through a second column stage.
Therefore, each cell token gets to revisit the context set after incorporating information from other features in the same row.
Only afterward does a separate readout compress the cell tokens into fixed-width row embeddings for in-context prediction.
This design preserves an additional round of within-row and cross-row interaction while retaining an efficient ICL backbone.

Moreover, \method{} accelerates inference by making within-row attention linear in the number of features.
While existing column-then-row architectures use full self-attention among cell tokens to form row embeddings, \method{} uses cross-attention through a fixed number of summary tokens in both row refinement and compression.
This replaces quadratic interactions among cell tokens with linear cost.

Pretrained entirely on approximately 36-million synthetic tables with varying sizes and causalities, \method{} demonstrates strong performance across three complementary benchmarks:

\begin{itemize}
\item \textbf{TabArena~\cite{erickson2026tabarena}.}
\method{} achieves 1785 Elo at a median inference time of 0.1043 seconds per 1K test samples.
It exceeds TabFM in Elo with 63.8 times faster inference.
It lies on the empirical Pareto frontier in both classification and regression.

\item \textbf{BeyondArena~\cite{purucker2026beyond}.}
Across a broader collection of datasets, \method{} ranks first among the evaluated models in both classification and regression, with Elo ratings of 1346 and 1394.
Its combined Elo of 1359 leads both foundation models and conventional supervised methods.

\item \textbf{ScoringBench~\cite{landsgesell2026scoringbench}.}
\method{} achieves Pareto-optimal performance across RMSE, $R^2$, MAE, and CRPS with a median inference time of approximately 0.3 seconds per fold.
It outperforms TabPFN-3.5-Fast on all four metrics while requiring only 40\% of the inference time of TabPFN-3.5.
\end{itemize}

We release the model and inference code to support reproducible comparisons and practical use.
\section{\method{}}
\label{sec:causilo}

\subsection{Refine First, Compress Later}
\label{sec:architecture}

\begin{figure}[!htb]
  \centering
  \includegraphics[width=\textwidth]{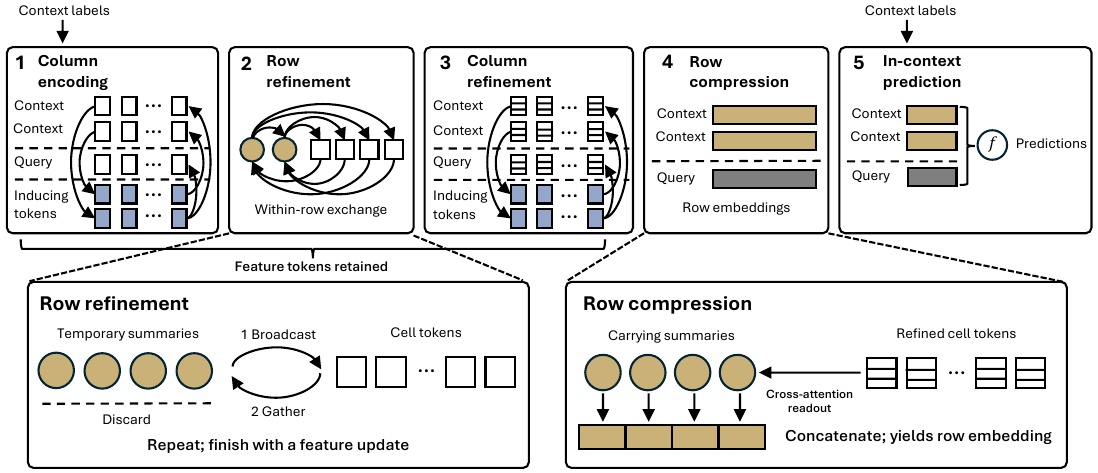}
  \caption{\textbf{Architecture of \method{}.} For a single input view, column encoding, row refinement, and column refinement update cell tokens while retaining the table's row and feature-group axes. Row compression then produces a fixed-width representation for each row, followed by dataset-level in-context prediction. The lower panels expose the architectural separation at the center of \method{}: temporary summaries mediate cell representation refinement, whereas an independent set of summaries performs the final readout.}
  \label{fig:architecture}
\end{figure}

A central architectural choice in \method{} is to refine cell representations before compressing them into row embeddings.
As illustrated in \Cref{fig:architecture}, the row- and column-refinement modules retain individual cell tokens.
The row-refinement module first exchanges information among cell tokens within each row.
The column-refinement module then revisits the context set using these enriched representations.

\mypar{From scalar cells to representations.}
\method{} receives labeled context rows and unlabeled query rows.
Retained features are partitioned into non-overlapping groups of three adjacent features, with zero padding for the final incomplete group.
Inspired by TabFM~\cite{google2026tabfm}, we encode observed scalars using sine and cosine features at 16 learned frequencies, followed by a shared linear projection to $\mathbb{R}^{d}$ with $d=128$.
Missing values use learned embeddings.
Both the frequencies and missing-value embeddings depend on the within-group position.
The three cell representations in each group are summed and scaled to form a single cell token.
Thus, each cell token represents one feature group in one row.
For context rows, a target embedding is added to every cell token.

\mypar{Column encoding.}
The column encoder updates cell tokens by exchanging information across rows within each column.\footnote{
After feature grouping, each column of the token grid corresponds to a group of adjacent input features.
We refer to these grouped positions as columns for convenience.
}
It uses the inducing-token attention pattern of Set Transformer~\cite{lee2019set}.
Let $H_{:j}\in\mathbb{R}^{N\times d}$ denote the cell tokens in column $j$ across $N$ rows, and let $H_{\mathcal C j}$ contain only the context rows.
At each layer, $M$ learned inducing tokens first \emph{gather} information from the context cell tokens.
Cell tokens in both context and query rows then read from the resulting summaries through a \emph{broadcast}:
\begin{equation}
\begin{aligned}
S_j
&=
\underbrace{
\mathcal{A}\!\left(U, H_{\mathcal C j}, H_{\mathcal C j}\right)
}_{\text{Gather}},
\quad
H'_{:j}
=
\underbrace{
\mathcal{A}\!\left(H_{:j}, S_j, S_j\right)
}_{\text{Broadcast}}.
\end{aligned}
\label{eq:column-encoding}
\end{equation}
Here, $U\in\mathbb{R}^{M\times d}$ denotes the learned inducing tokens,
$S_j\in\mathbb{R}^{M\times d}$ the resulting column summaries, and
$\mathcal{A}(Q,K,V)$ an attention update.
This enables context-to-context and context-to-query information exchange within each column.
Note that query cell tokens do not contribute to the shared summaries.

\mypar{Row refinement.}
The column encoder updates cell tokens across rows, but processes each column independently.
\method{} therefore introduces within-row interactions before compression.
For row $i$, let $H_i\in\mathbb{R}^{G\times d}$ denote its $G$ cell tokens and
$S_i\in\mathbb{R}^{K\times d}$ a set of $K=4$ temporary summary tokens.
Each refinement round alternates a \emph{broadcast} from the summaries to the cell tokens and a \emph{gather} of the updated cell representations back into the summaries:
\begin{equation}
\begin{aligned}
H_i'
&=
\underbrace{
\mathcal{A}\!\left(H_i,S_i,S_i\right)
}_{\text{Broadcast}},
\qquad
S_i'
=
\underbrace{
\mathcal{A}\!\left(S_i,H_i',H_i'\right)
}_{\text{Gather}}.
\end{aligned}
\label{eq:row-refinement}
\end{equation}
The summary tokens carry information across refinement layers, collecting information from cell tokens and redistributing it within the row.
After the final broadcast, the summaries are discarded and the refined cell tokens pass to the next stage.

\mypar{Column refinement.}
After row refinement, each cell token carries information about other features in the same row.
\method{} applies a second column stage so that these enriched representations can revisit the context set before compression.
Using the row-refined cell tokens $\bar H$, the model again performs a context-only \emph{gather} followed by an all-row \emph{broadcast}:
\begin{equation}
\begin{aligned}
\bar S_j
&=
\underbrace{
\bar{\mathcal{A}}\!\left(
\bar U,
\bar H_{\mathcal C j},
\bar H_{\mathcal C j}
\right)
}_{\text{Gather}},
\qquad
H^{\star}_{:j}
=
\underbrace{
\bar{\mathcal{A}}\!\left(
\bar H_{:j},
\bar S_j,
\bar S_j
\right)
}_{\text{Broadcast}}.
\end{aligned}
\label{eq:column-refinement}
\end{equation}
This stage follows the same attention pattern as the first column encoder but uses independent parameters.
Because its input cell tokens already contain within-row information, the second column stage can use cross-feature evidence when updating representations across rows.

\mypar{Row compression.}
After refinement, \method{} compresses each row using a separate set of $K=4$ learned summary tokens.
Let $P_i^{(\ell)}$ denote the summary tokens at layer $\ell$ and $H_i^\star$ the refined cell tokens for row $i$.
The attention update is
\begin{equation}
P_i^{(\ell+1)}
=
\mathcal{A}\!\left(
P_i^{(\ell)},
[H_i^\star; P_i^{(\ell)}],
[H_i^\star; P_i^{(\ell)}]
\right),
\label{eq:row-readout}
\end{equation}
where $[\cdot;\cdot]$ denotes concatenation along the token dimension.
Unlike row refinement, this module updates only the summary tokens; the cell tokens remain fixed.
The final summaries are individually RMS-normalized~\cite{zhang2019rmsnorm} and concatenated into a $Kd=512$-dimensional row embedding.

\mypar{In-context prediction.}
A separate Transformer~\cite{vaswani2017attention} processes the row embeddings to predict query targets from the labeled context.
Context targets are embedded again and added to their corresponding row embeddings.
Let $U_{\mathcal C}^{(\ell)}$ and $U_{\mathcal Q}^{(\ell)}$ denote the context and query row representations at layer $\ell$.
At each layer, context representations attend to one another, while query representations attend only to the context:
\begin{equation}
\begin{aligned}
U_{\mathcal C}^{(\ell+1)}
&=
\mathcal{A}\!\left(
U_{\mathcal C}^{(\ell)},
U_{\mathcal C}^{(\ell)},
U_{\mathcal C}^{(\ell)}
\right),
\quad
U_{\mathcal Q}^{(\ell+1)}
=
\mathcal{A}\!\left(
U_{\mathcal Q}^{(\ell)},
U_{\mathcal C}^{(\ell)},
U_{\mathcal C}^{(\ell)}
\right).
\end{aligned}
\label{eq:icl-attention}
\end{equation}
After $L=12$ Transformer layers, a two-layer head with GELU~\cite{hendrycks2016gaussian} maps each query row representation to class logits or regression outputs.
Across the architecture, attention blocks use pre-RMS normalization, residual connections, and SwiGLU~\cite{shazeer2020glu} feedforward layers.

\subsection{Linear Feature Scaling by Construction}
\label{sec:fast-inference}
\Needspace{0.5\baselineskip}
\begin{wrapfigure}{r}{0.6\textwidth}
  \centering
  \vspace{-01.2em}
  \includegraphics[width=\linewidth]{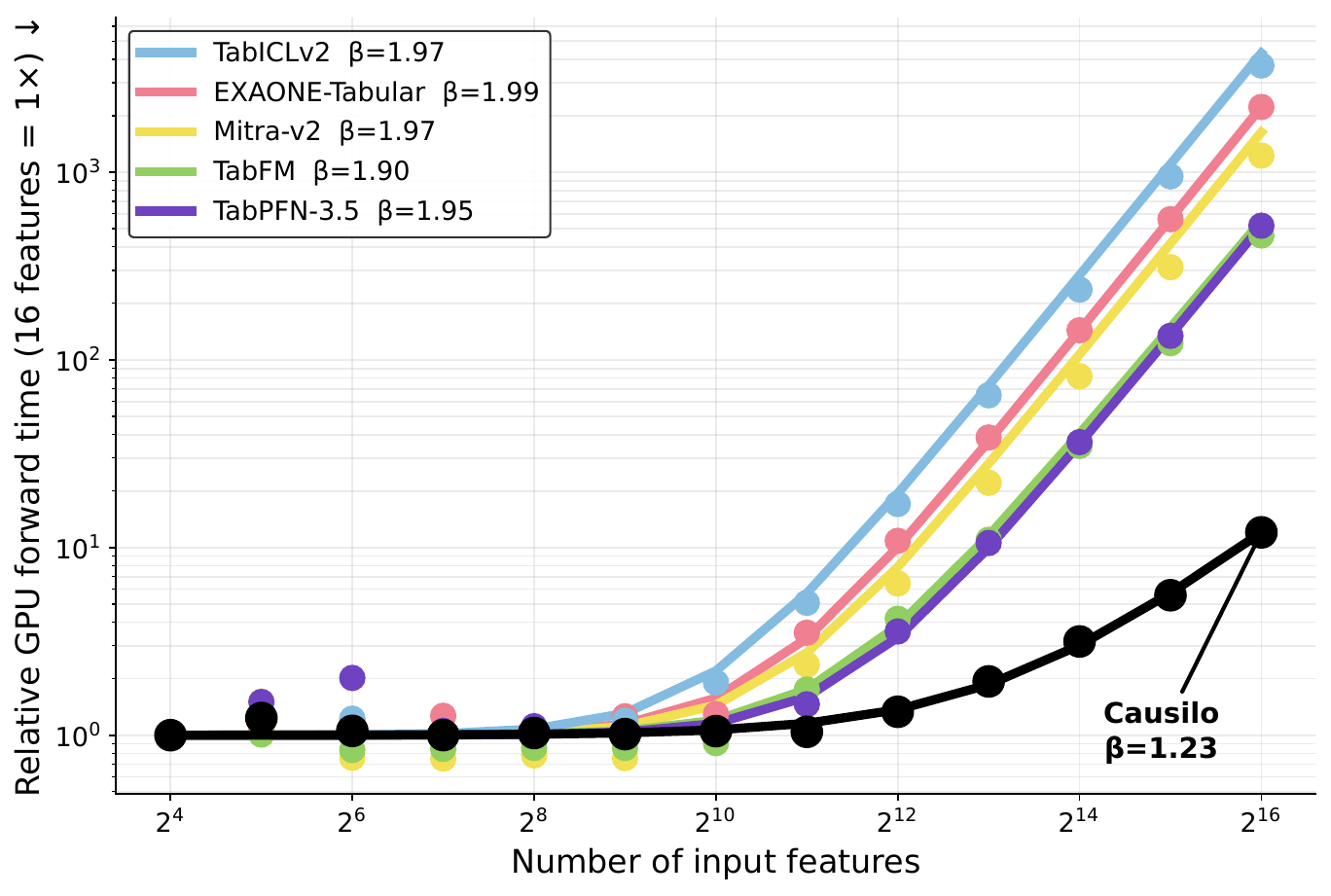}
  \caption{\textbf{Feature-wise forward-time scaling of tabular foundation models.}
  Markers show the median of 50 runs, normalized by the measured time at 16 features. Lines show fits of $T(F)=c+aF^\beta$, normalized by the fitted time at 16 features; the annotated $\beta$ values are estimated over the displayed feature range.}
   \vspace{-01.2em}
  \label{fig:feature-scaling}
\end{wrapfigure}

\mypar{Cross-attention throughout both row stages.}
Existing frontier \acp{tfm} retain feature-side self-attention while individual cell representations remain explicit.
\method{} instead routes within-row communication through a fixed set of summaries throughout both row modules.
Refinement exchanges information in both directions between cell tokens and summary tokens; compression updates only the summaries.
This asymmetry eliminates quadratic attention within rows.

\begin{table}[t]
\centering
\caption{Attention complexity per row and per head for $G$ cell tokens and $K$ summary tokens. With fixed $K$, all three operations in \method{} scale linearly with $G$.}
\label{tab:attention-complexity}
\begin{tabular}{@{}ll@{}}
\toprule
Operation & Complexity \\
\midrule
Full self-attention over cell and summary tokens
& $\mathcal{O}((G+K)^2)$ \\
\method{}: one refinement round
& $\mathcal{O}(K(G+K))$ \\
\method{}: final row-refinement broadcast
& $\mathcal{O}(GK)$ \\
\method{}: one compression layer
& $\mathcal{O}(K(G+K))$ \\
\bottomrule
\end{tabular}
\end{table}
\mypar{Cost as the feature count grows.}
Table~\ref{tab:attention-complexity} summarizes the attention cost per row and per head for $G$ cell tokens and $K$ summary tokens.
With fixed $K$, attention in both row modules scales linearly with $G$, whereas full self-attention scales quadratically.
Both column stages  use a fixed number of inducing tokens, keeping their attention cost linear in the number of rows and feature.
Full attention across context rows appears only in the ICL Transformer.
For $N_s$ context rows and $N_q$ query rows, the attention cost of a forward pass is therefore  $\mathcal{O}\!\left(
    (N_s+N_q)G + N_s(N_s+N_q)
  \right)$,
with widths, depths, and summary counts held fixed.
The first term covers cell-level column and row processing, whereas second term covers the attention in the ICL Transformer.

\mypar{Scaling in practice.}
We use a targeted GPU microbenchmark to examine whether feature-interaction time grows linearly or quadratically with feature-axis input size.
For each pretrained model---\method{}, TabFM~\cite{google2026tabfm}, Mitra-v2~\cite{tao2026mitra}, TabPFN-3.5~\cite{priorlabs2026tabpfn35}, TabICLv2~\cite{qu2026tabiclv2}, and EXAONE-Tabular~\cite{eo2026exaone}---we fix the input at 16 rows and vary the feature-size parameter $F$ from $2^4$ to $2^{16}$.
We time the first row-mixing block of \method{}, the attention submodules of TabFM's first row-interaction stage, and the feature-attention submodules across all layers of Mitra-v2 and EXAONE-Tabular.
For TabICLv2, we time the embedding-aggregation routine within the row-interaction module; for TabPFN-3.5, we time the complete column-aggregation module.
All these operations mix feature tokens within each row.
After model setup, we run the selected operation 20 times for warm-up and record 50 CUDA-event timings.

Markers in Figure~\ref{fig:feature-scaling} show median timings normalized by each model's measured median at $F=16$. We fit the unnormalized medians to $T(F)=c+aF^\beta$, where $c\geq0$ captures approximately fixed overhead, $a$ sets the runtime scale, and $\beta$ describes feature-dependent growth.
Exponents of $\beta=1$ and $\beta=2$ correspond to linear and quadratic growth, respectively.
The fitted curves are then normalized by their values at $F=16$, with exponents estimated over the full displayed range.
\method{} yields $\beta=1.23$, close to linear scaling, whereas the other models yield $\beta=1.90$--$1.99$, close to quadratic scaling.
This agrees with the expected benefit of replacing full self-attention among cell tokens within each row with cross-attention through a fixed number of summary tokens.

\subsection{Bounded QASSMax}
\label{sec:stable-qassmax}

In attention over tabular data, keys can represent context rows within a column or cell tokens within a row, so context length varies with dataset size and table width.
As the context grows, softmax can spread probability mass across more entries and dilute attention to relevant ones.
Scalable-Softmax (SSMax)~\cite{nakanishi2025scalable} addresses this effect with length-dependent scaling.
TabICLv2~\cite{qu2026tabiclv2} introduces query-aware scalable softmax (QASSMax), which also adapts the scale to query content.
\method{} retains this adaptivity while explicitly bounding the scaling factors.

\mypar{The unbounded formulation.}
Let $q_h$ denote the query vector for head $h$, $i$ a coordinate within that head, and $n$ the number of keys.
TabICLv2 rescales each query coordinate as
\begin{equation}
\widetilde q_{hi}^{\mathrm{TabICLv2}}
=
q_{hi}\,
\underbrace{f_{hi}(\log n)}_{\text{length-dependent scale}}
\underbrace{\bigl[1+\tanh(g_i(q_h))\bigr]}_{\text{query modulation}}.
\label{eq:qass-tabicl}
\end{equation}
The network $f$ produces a length-dependent scale for each head and coordinate, while $g$ modulates it for the current query.
Although the query-dependent factor lies in $(0,2)$, the output of $f$ is unconstrained.
The resulting multiplier can therefore be negative or arbitrarily large.

\mypar{Positive and bounded scaling.}
\method{} retains the length-dependent network $f$ and query-dependent network $g$, but bounds their outputs before applying them to the query.
For an output $z$ from either network, we use
\begin{equation}
\operatorname{Bound}_L(z)
=
\exp\!\left[(\log L)\tanh\!\left(\frac{z}{\log L}\right)\right],
\qquad L>1.
\label{eq:qass-bound}
\end{equation}
This produces a positive multiplier between $1/L$ and $L$, with zero mapped to one.
We set $L=8$ for $f$ and $L=2$ for $g$. To accommodate context lengths that may differ from those seen during pretraining, \method{} also introduces a smaller length-dependent network $m$.
It provides a bounded correction to the scale produced by $f$, independently of query content.
The complete update is
\begin{equation}
\widetilde q_{hi}
=
q_{hi}\,
\underbrace{\operatorname{Bound}_8(f_{hi}(\log n))}_{\text{primary length scale}}
\underbrace{\operatorname{Bound}_2(m_{hi}(\log n))}_{\text{length correction}}
\underbrace{\operatorname{Bound}_2(g_i(q_h))}_{\text{query modulation}}.
\label{eq:qass-causilo}
\end{equation}
The correction can increase or decrease the primary scale by up to a factor of two.
The total multiplier remains between $1/32$ and $32$, preserving the sign of each query coordinate and preventing unbounded rescaling.
The networks $f$ and $g$ are two-layer MLPs with 64 hidden units and GELU activations.
The correction network $m$ uses 16 hidden units and a $\tanh$ activation.

\subsection{Inference-Time Ensembling}
\label{sec:estimator-ensemble}

Inference-time ensembling is widely used in \acp{tfm}~\cite{eo2026exaone, grinsztajn2026tabpfn, google2026tabfm}.
\method{} constructs ensemble members by varying feature normalization,
feature order, and, for classification, class labels.
All members use the same weights, and their predictions
are combined after restoring the original class order or target scale.

\mypar{Normalization schemes.}
Ensemble members cycle through four normalization schemes:
standardization, rank-to-Gaussian transformation, robust scaling,
and power transformation.
Each scheme is fitted on the context set and followed by smooth
tail compression, which reduces the magnitude of extreme values
while leaving values within context-derived bounds unchanged.
The same fitted transformation is applied to query rows without
re-estimating its parameters.

\mypar{Feature permutations.}
\method{} combines three adjacent features into a single token, therefore
feature order determines which features are encoded together.
We vary this order across ensemble members to provide different
feature groupings.
The first member preserves the original feature order.
For the remaining members, we randomly arrange the features on
a shared random ring, then generate permutations
by varying the traversal stride and starting position.
Candidate strides are coprime to the feature count so that
each traversal visits every feature exactly once.
Among these candidates, we favor strides with less frequently
used one- and two-step circular distances to discourage
repeated feature combinations.

\mypar{Prediction aggregation.}
For classification, we permute class labels across ensemble members
using cyclic shifts of a random class ordering.
Over each complete cycle of $C$ members, each of the $C$ classes
occupies every class index once.
We restore each member's logits to the original class order,
average them across members, and apply softmax.
For regression, we average the output channels of each member
to obtain a point prediction.
We then transform these predictions back to the original target
scale and average them across members.

\Needspace{5\baselineskip}
\FloatBarrier
\section{Evaluation}
\label{sec:evaluation}

We evaluate \method{} on three complementary benchmarks of real-world tabular prediction. TabArena~\cite{erickson2026tabarena} assesses classification and regression performance against strong foundation and supervised models. BeyondArena~\cite{purucker2026beyond} extends this comparison to diverse data regimes, while ScoringBench~\cite{landsgesell2026scoringbench} evaluates both point predictions and predictive distributions for regression. We also examine inference latency on TabArena and ScoringBench to assess where \method{} lies on the performance--efficiency frontier.

\subsection{TabArena}
\label{sec:tabarena}
\acreset{tfm}

\begin{figure}[!htbp]
  \centering
  \includegraphics[draft=false,width=\linewidth]{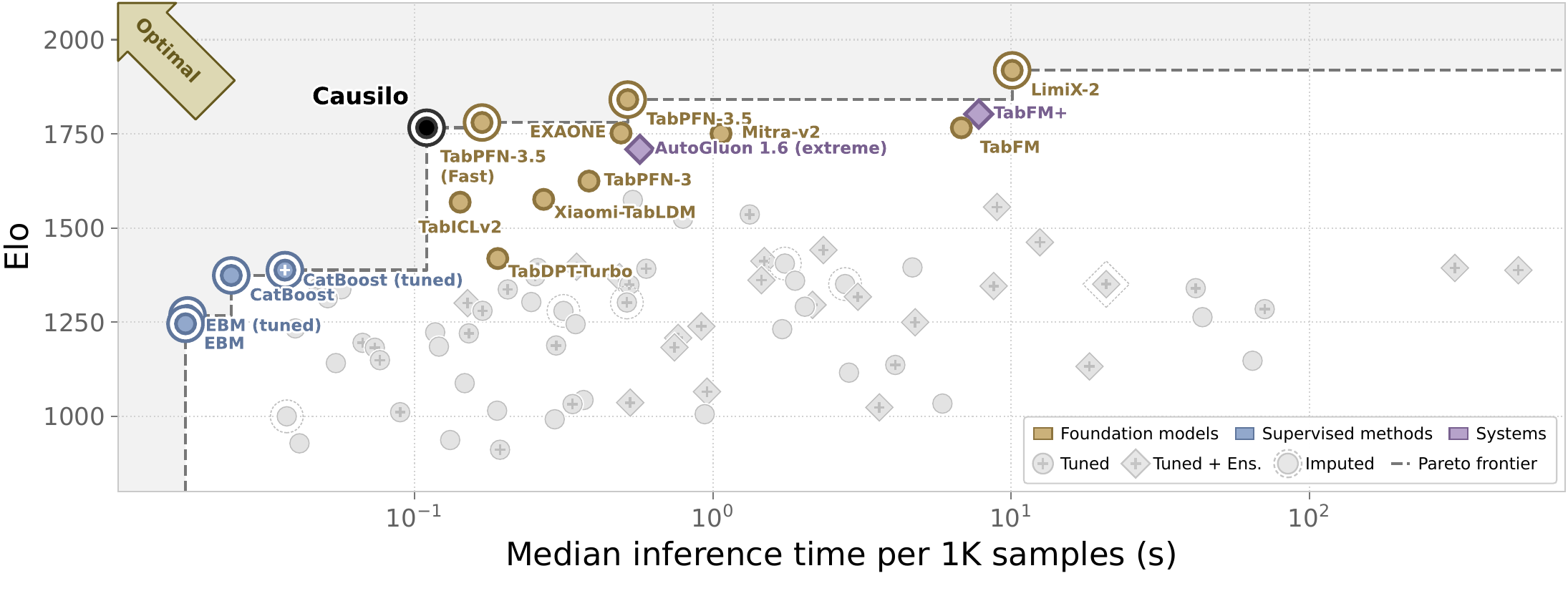}
  \caption{\textbf{TabArena classification performance--efficiency frontier.} Higher Elo and lower median inference time per 1K test samples are better; the time axis is logarithmic. The dashed line traces the Pareto frontier among the plotted configurations. \method{} achieves 1766.3 Elo at 0.1096 seconds per 1K test samples.}
  \label{fig:tabarena-classification}
\end{figure}

TabArena is a continuously maintained benchmark for supervised tabular learning, comprising 51 real-world datasets---38 for classification and 13 for regression. Its standardized evaluation pipeline uses shared data splits to compare tree ensembles, neural networks, and foundation models, with support for default configurations, hyperparameter tuning, and ensembling. Predictive performance is measured by ROC--AUC for binary classification, log loss for multiclass classification, and RMSE for regression. Elo ratings summarize pairwise comparisons across datasets, with higher ratings indicating stronger aggregate performance. We use the latest public results as of 18 September 2026 and Elo ratings are computed over the full leaderboard, including system submissions such as AutoGluon~\cite{erickson2020autogluon} and TabFM+. A configuration lies on the Pareto frontier when no other configuration achieves both at least as high an Elo and at least as low an inference time, with a strict improvement in one dimension.

\mypar{Classification.}\label{sec:tabarena-classification}
\method{} achieves 1766 Elo at 0.1096 seconds per 1K test samples (Figure~\ref{fig:tabarena-classification}). It essentially matches TabFM's 1766 Elo with a 62.1-fold speedup over its 6.8094-second inference time. Relative to TabICLv2, it gains 198 Elo while requiring 22.8\% less inference time. It also exceeds EXAONE-Tabular and Mitra-v2 in Elo with 4.49-fold and 9.71-fold speedups, respectively. These results place \method{} on the Pareto frontier, delivering competitive predictive performance at approximately one-tenth of a second per 1K test samples.

Higher-Elo configurations extend the frontier at increasing inference cost. TabPFN-3.5 reaches 1841 Elo at 0.5178 seconds, while LimiX-2~\cite{zhang2026limix2contextualmechanismnetwork} leads with 1919 Elo but requires 10.0810 seconds, approximately 92 times \method{}'s inference time. System submissions do not improve this frontier: TabFM+ reaches 1803 Elo at 7.7865 seconds and is dominated by TabPFN-3.5. 

\begin{figure}[!htbp]
  \centering
  \includegraphics[draft=false,width=\linewidth]{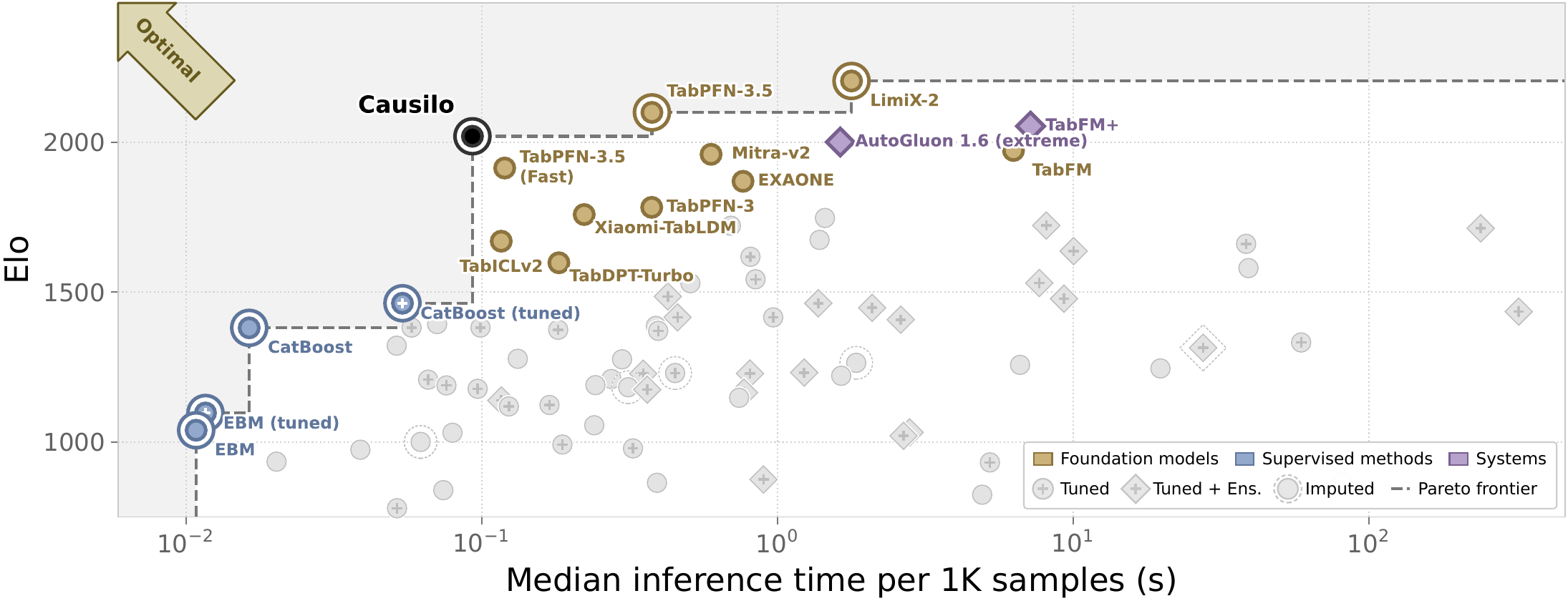}
  \caption{\textbf{TabArena regression performance--efficiency frontier.} Higher Elo and lower median inference time per 1K test samples are better; the time axis is logarithmic. The dashed line traces the Pareto frontier among the plotted configurations. \method{} achieves 2019.3 Elo at 0.0930 seconds per 1K test samples.}
  \label{fig:tabarena-regression}
\end{figure}

\mypar{Regression.}\label{sec:tabarena-regression}
\method{} reaches 2019 Elo at 0.0930 seconds per 1K test samples (Figure~\ref{fig:tabarena-regression}). Its advantage over TabFM is larger than in classification, as it gains 47 Elo while running 67.6 times faster. It also exceeds Mitra-v2 by 60 Elo with 6.41 times faster inference. Against TabICLv2, \method{} gains 350 Elo and reduces inference time by 20.1\%. Thus, its low latency is accompanied by a clear improvement in aggregate predictive performance over these baselines.

Unlike in classification, \method{} dominates TabPFN-3.5-Fast, gaining 106 Elo with 22.2\% less inference time. The high-Elo frontier therefore connects \method{} directly to TabPFN-3.5, which reaches 2099 Elo at 0.3765 seconds, and then to LimiX-2, which reaches 2204 Elo at 1.7817 seconds. These models require 4.05 and 19.1 times \method{}'s inference time, respectively. TabFM+ and AutoGluon 1.6 noncommercial score above \method{}, but both are dominated by TabPFN-3.5.

\mypar{Combined.}\label{sec:tabarena-combined}
Figure~\ref{fig:overview} summarizes the combined comparison. \method{} achieves 1785 Elo at 0.1043 seconds per 1K test samples and occupies the low-latency end of the high-performance frontier. TabPFN-3.5 raises Elo to 1861 at 0.4727 seconds and LimiX-2 attains the highest combined Elo of 1943 at 9.0278 seconds, requiring 86.6 times \method{}'s inference time. Therefore, \method{} is a compelling choice when both predictive accuracy and fast inference are essential.

\FloatBarrier
\subsection{BeyondArena}
\label{sec:beyondarena}
\acreset{tfm}
\begin{figure}[t]
  \centering
  \begin{subfigure}[t]{0.485\textwidth}
    \centering
    \includegraphics[draft=false,width=\linewidth]{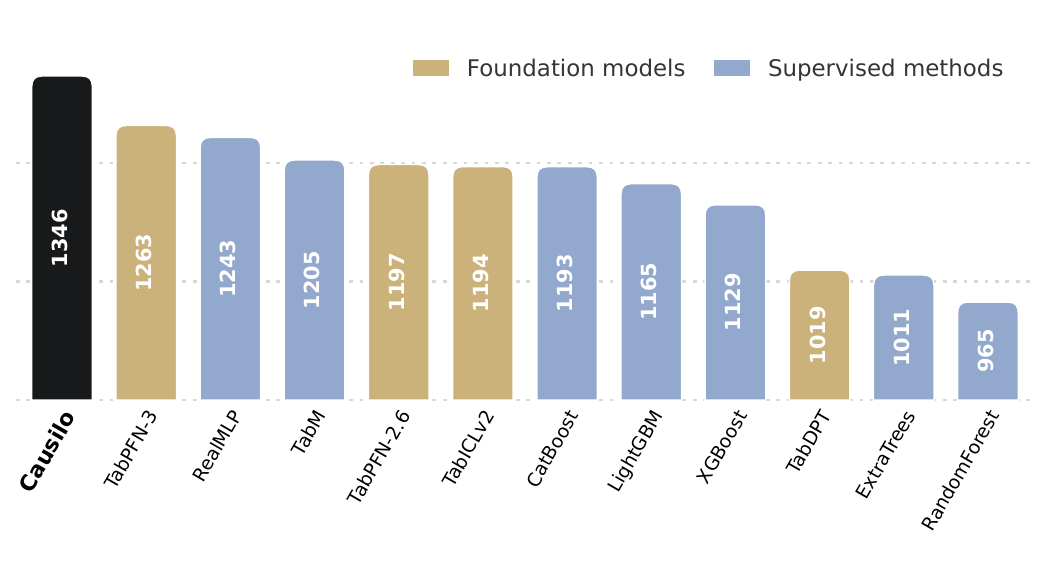}
    \caption{Classification.}
    \label{fig:beyondarena-classification}
  \end{subfigure}\hfill
  \begin{subfigure}[t]{0.485\textwidth}
    \centering
    \includegraphics[draft=false,width=\linewidth]{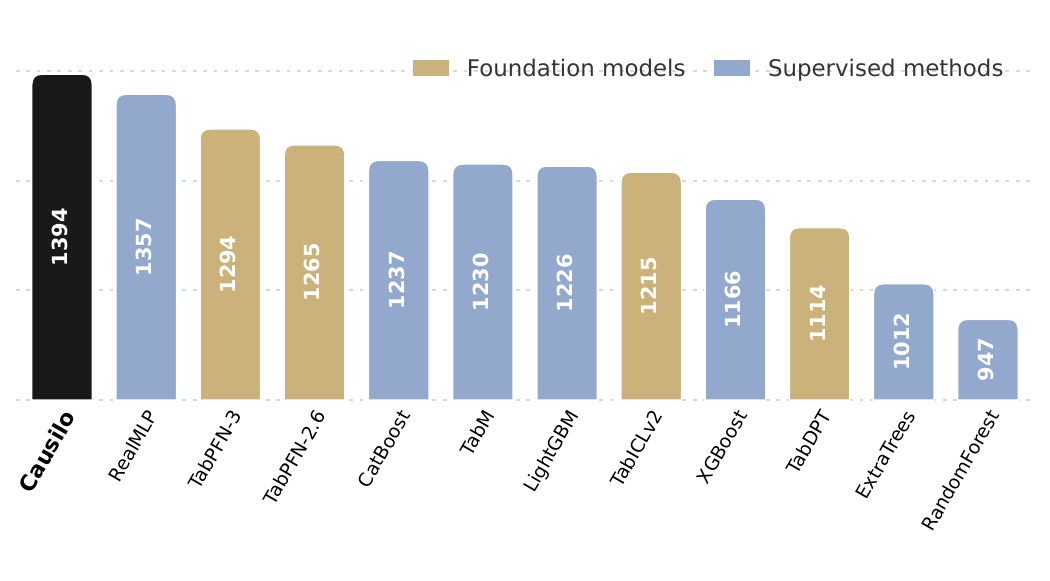}
    \caption{Regression.}
    \label{fig:beyondarena-regression}
  \end{subfigure}
  \caption{BeyondArena Elo of the top 12 models for classification and regression, excluding system submissions and retaining only the highest-rated variant from each model family within each comparison. Higher is better. Black denotes \method{}, gold denotes other foundation models, and blue denotes supervised methods.}
  \label{fig:beyondarena-results}
\end{figure}
BeyondArena~\cite{purucker2026beyond} tests whether strong tabular prediction extends beyond standard IID settings. Its 142 curated datasets span IID, temporal, and grouped splits, a wide range of sample sizes and feature counts, and tables with text or high-cardinality categorical features. These settings test generalization under changes that random train--test splits can overlook. The benchmark therefore complements TabArena by covering data regimes in which conventional supervised models remain competitive. The latest public BeyondArena release includes models up to TabPFN-3. We therefore use its published baseline results and add only \method{}, which we evaluate locally. 

\mypar{Classification.}\label{sec:beyond-classification}
\method{} leads the classification comparison with 1346 Elo (Figure~\ref{fig:beyondarena-classification}), ahead of TabPFN-3 at 1263 and RealMLP~\cite{holzmuller2024better} at 1243 by 83 and 103 points, respectively. TabM~\cite{gorishniy2025tabm}, TabPFN-2.6~\cite{priorlabs2026tabpfn26}, TabICLv2, and CatBoost~\cite{prokhorenkova2018catboost} achieve similar ratings of 1205, 1197, 1194, and 1193. Conventional supervised models remain competitive: RealMLP outperforms all evaluated TFMs except \method{} and TabPFN-3, while TabM surpasses TabPFN-2.6 and TabICLv2. \method{} leads both foundation models and conventional supervised methods across this broader collection of classification tasks.

\mypar{Regression.}\label{sec:beyond-regression}
\method{} also ranks first in regression with 1394 Elo (Figure~\ref{fig:beyondarena-regression}). Here, RealMLP is the closest competitor at 1357, narrowing the lead to 37 points. TabPFN-3 follows at 1294 and TabPFN-2.6 at 1265, trailing \method{} by 100 and 129 points. \method{} retains the highest aggregate rating despite the change in its closest competitor between classification and regression.

\mypar{Combined.}\label{sec:beyond-combined}
Across both tasks, \method{} achieves the highest combined Elo of 1359 (Table~\ref{tab:beyondarena-combined}). RealMLP and TabPFN-3 are nearly tied at 1274 and 1270, leaving margins of 85 and 89 points. \method{} also attains the lowest improvability (6.4\%) and average rank (6.2), together with the highest aggregated win count (47.3). Together with the task-specific rankings, these results show that \method{}'s aggregate advantage extends beyond TabArena setting and holds against both foundation and supervised models.

\begin{table}[t]
  \centering
  \caption{Combined BeyondArena results for 29 evaluated configurations, sorted by Elo. D denotes default, T denotes tuned, and T+E denotes tuned and ensembled configurations. Elo subscripts give the reported lower and upper confidence deviations. Wins are aggregated first-place counts and may be fractional. Best values are bold, and the \method{} row is shaded.}
  \begingroup
  \scriptsize
  \setlength{\tabcolsep}{1.5pt}
  \renewcommand{\arraystretch}{1.12}
  \begin{tabular*}{\textwidth}{@{\extracolsep{\fill}}l c c c c !{\hspace{5pt}\vrule\hspace{5pt}} l c c c c@{}}
    \toprule
    \textbf{Model} & \shortstack{\textbf{Elo}\\$\uparrow$} & \shortstack{\textbf{Improv-}\\\textbf{ability} $\downarrow$} & \shortstack{\textbf{Avg.}\\\textbf{rank} $\downarrow$} & \shortstack{\textbf{\#wins}\\$\uparrow$} & \textbf{Model} & \shortstack{\textbf{Elo}\\$\uparrow$} & \shortstack{\textbf{Improv-}\\\textbf{ability} $\downarrow$} & \shortstack{\textbf{Avg.}\\\textbf{rank} $\downarrow$} & \shortstack{\textbf{\#wins}\\$\uparrow$} \\
    \midrule
    \cellcolor{numslight}\textbf{\method{}} & \cellcolor{numslight}$\mathbf{1359}_{-40,+45}$ & \cellcolor{numslight}\textbf{6.4\%} & \cellcolor{numslight}\textbf{6.2} & \cellcolor{numslight}\textbf{47.3} & TabM (D) & $1100_{-27,+26}$ & 15.5\% & 14.4 & 1.8 \\
    RealMLP (T+E) & $1274_{-26,+30}$ & 11.9\% & 8.5 & 7.1 & RealMLP (D) & $1053_{-27,+22}$ & 17.4\% & 16.1 & 3.0 \\
    TabPFN-3 (D) & $1270_{-41,+43}$ & 10.0\% & 8.6 & 12.5 & TabDPT (D) & $1047_{-47,+37}$ & 19.0\% & 16.3 & 5.5 \\
    TabPFN-2.6 (D) & $1216_{-37,+40}$ & 12.4\% & 10.4 & 9.4 & ExtraTrees (T+E) & $1011_{-30,+22}$ & 19.6\% & 17.5 & 1.0 \\
    TabM (T+E) & $1211_{-27,+29}$ & 13.2\% & 10.5 & 3.2 & XGBoost (D) & $1000_{-26,+31}$ & 19.0\% & 17.9 & 1.0 \\
    CatBoost (T+E) & $1205_{-22,+27}$ & 13.6\% & 10.7 & 4.8 & LightGBM (D) & $988_{-21,+26}$ & 19.1\% & 18.3 & 0.8 \\
    RealMLP (T) & $1202_{-24,+31}$ & 13.4\% & 10.8 & 6.6 & ExtraTrees (T) & $977_{-34,+28}$ & 20.3\% & 18.7 & 0.5 \\
    TabICLv2 (D) & $1198_{-41,+42}$ & 11.9\% & 10.9 & 8.2 & RandomForest (T+E) & $961_{-29,+24}$ & 20.7\% & 19.3 & 0.9 \\
    CatBoost (T) & $1189_{-24,+26}$ & 13.8\% & 11.2 & 5.6 & RandomForest (T) & $938_{-32,+25}$ & 21.3\% & 20.0 & 0.7 \\
    LightGBM (T+E) & $1182_{-23,+28}$ & 14.4\% & 11.5 & 3.4 & LinearModel (T+E) & $880_{-39,+30}$ & 27.5\% & 21.7 & 2.8 \\
    TabM (T) & $1171_{-28,+27}$ & 13.9\% & 11.9 & 3.8 & RandomForest (D) & $874_{-35,+31}$ & 23.4\% & 21.9 & 0.8 \\
    LightGBM (T) & $1144_{-23,+28}$ & 14.9\% & 12.8 & 1.7 & ExtraTrees (D) & $865_{-42,+38}$ & 24.1\% & 22.2 & 1.2 \\
    XGBoost (T+E) & $1139_{-25,+32}$ & 16.0\% & 13.0 & 1.8 & LinearModel (T) & $850_{-45,+33}$ & 28.3\% & 22.6 & 2.3 \\
    CatBoost (D) & $1132_{-21,+31}$ & 15.4\% & 13.2 & 1.1 & LinearModel (D) & $796_{-52,+40}$ & 30.7\% & 23.9 & 0.5 \\
    XGBoost (T) & $1117_{-25,+29}$ & 16.2\% & 13.8 & 2.7 &  &  &  &  &  \\
    \bottomrule
  \end{tabular*}
  \endgroup
  \label{tab:beyondarena-combined}
\end{table}

\begin{figure}[t]
  \centering
  \includegraphics[draft=false,width=\textwidth]{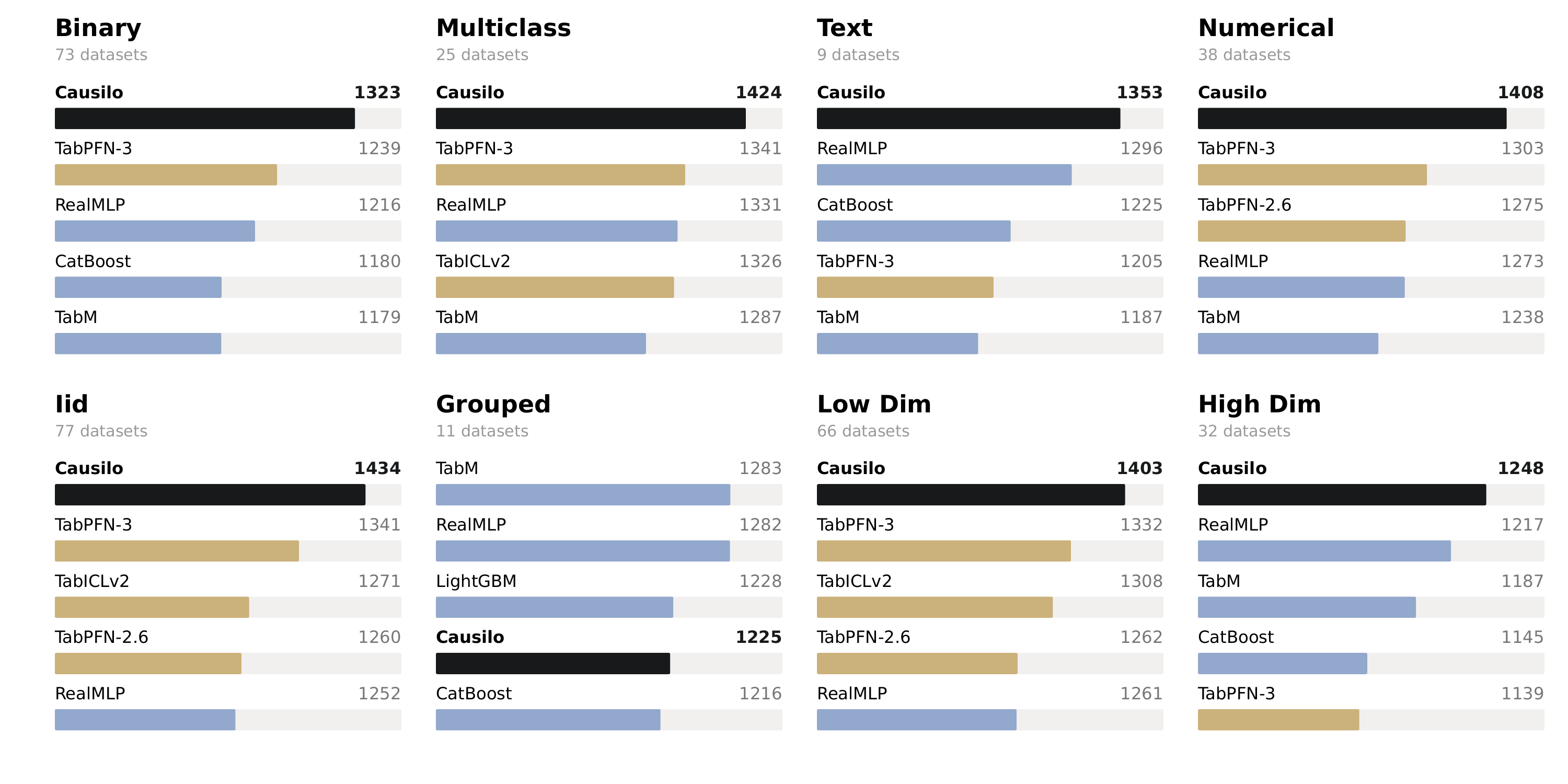}
  \caption{BeyondArena classification Elo by subgroup. Each panel shows the top five model families, retaining the highest-rated variant within each family. Dataset counts appear below the subgroup names. Black denotes \method{}, gold other TFMs, and blue conventional supervised models. Higher Elo is better.}
  \label{fig:beyondarena-subgroup-classification}
\end{figure}

\mypar{Classification subgroups.}
\method{} leads both binary and multiclass classification with 1323 and 1424 Elo, respectively (Figure~\ref{fig:beyondarena-subgroup-classification}). TabPFN-3 is the nearest competitor in both, trailing by 84 and 83 points. The nearly identical margins show that the classification advantage holds across both target types. TabPFN-3 is also the closest competitor on IID and numerical datasets, where \method{} leads by 93 points (1434 versus 1341) and 105 points (1408 versus 1303).

The competing models change as the feature composition changes. On low-dimensional tables, with at most 100 features after preprocessing, \method{} leads TabPFN-3 by 71 points (1403 versus 1332). On high-dimensional tables, RealMLP becomes the nearest competitor, narrowing the margin to 31 points (1248 versus 1217); TabM and CatBoost also rank above TabPFN-3. Grouped classification presents a different ordering. Across 11 datasets, \method{} is the highest-ranked TFM at 1225 Elo, behind TabM at 1283, RealMLP at 1282, and LightGBM~\cite{ke2017lightgbm} at 1228, but ahead of CatBoost at 1216. 

\begin{figure}[t]
  \centering
  \includegraphics[draft=false,width=\textwidth]{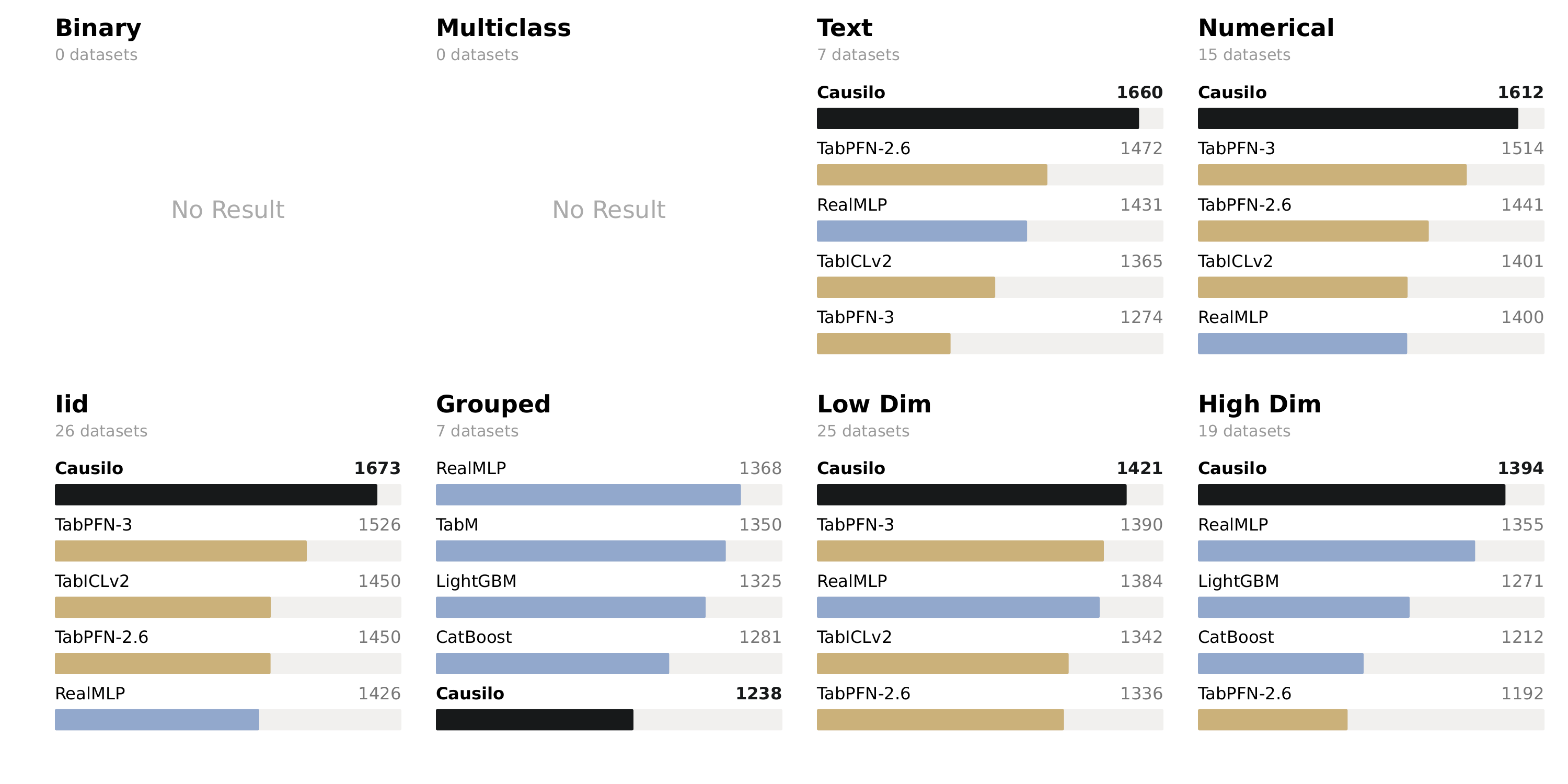}
  \caption{BeyondArena regression Elo by subgroup, using the same model-family selection and colors as Figure~\ref{fig:beyondarena-subgroup-classification}. \method{} leads five of the six displayed subgroups and is the highest-ranked TFM on grouped regression. Higher Elo is better.}
  \label{fig:beyondarena-subgroup-regression}
\end{figure}

\mypar{Regression subgroups.}
\method{} leads on IID, numerical, and text-containing datasets, exceeding the strongest baseline in each by 147, 98, and 188 Elo, respectively (Figure~\ref{fig:beyondarena-subgroup-regression}). Its advantage is smaller across dimensionality subsets; i.e., it leads TabPFN-3 by 31 Elo on low-dimensional tables.

Grouped regression is more challenging. Across seven datasets, \method{} remains the strongest TFM at 1238 Elo but trails RealMLP by 130 points and also ranks below TabM, LightGBM, and CatBoost. These results show broad strength across data regimes, while identifying grouped generalization as a remaining gap relative to conventional supervised methods.

\subsection{ScoringBench}
\label{sec:scoringbench}
\acreset{tfm}

\begin{figure}[t]
  \centering
  \begin{subfigure}[t]{0.485\textwidth}
    \centering
    \includegraphics[draft=false,width=\linewidth]{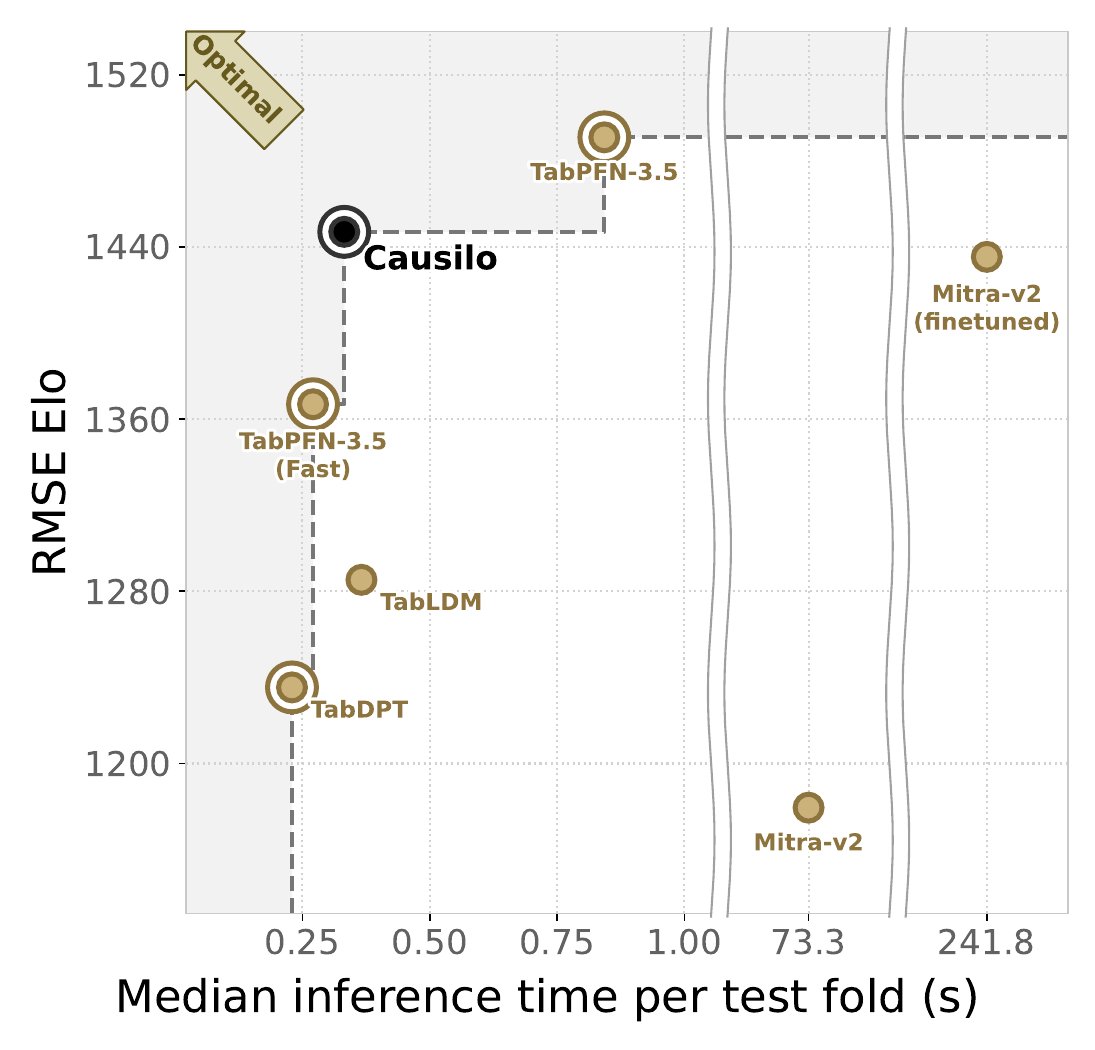}
    \caption{RMSE.}
    \label{fig:scoringbench-rmse}
  \end{subfigure}\hfill
  \begin{subfigure}[t]{0.485\textwidth}
    \centering
    \includegraphics[draft=false,width=\linewidth]{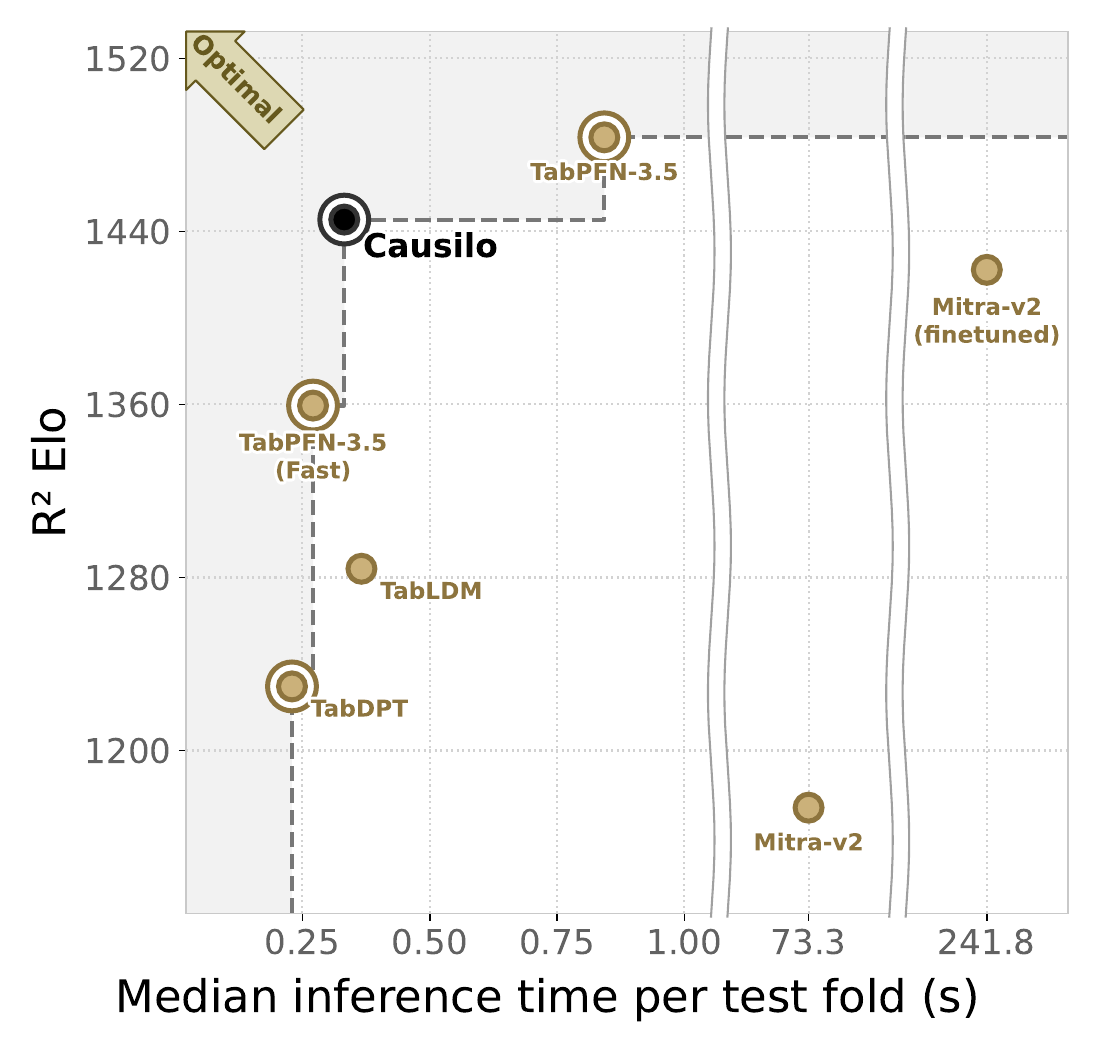}
    \caption{$R^2$.}
    \label{fig:scoringbench-r2}
  \end{subfigure}

  \vspace{0.6em}

  \begin{subfigure}[t]{0.485\textwidth}
    \centering
    \includegraphics[draft=false,width=\linewidth]{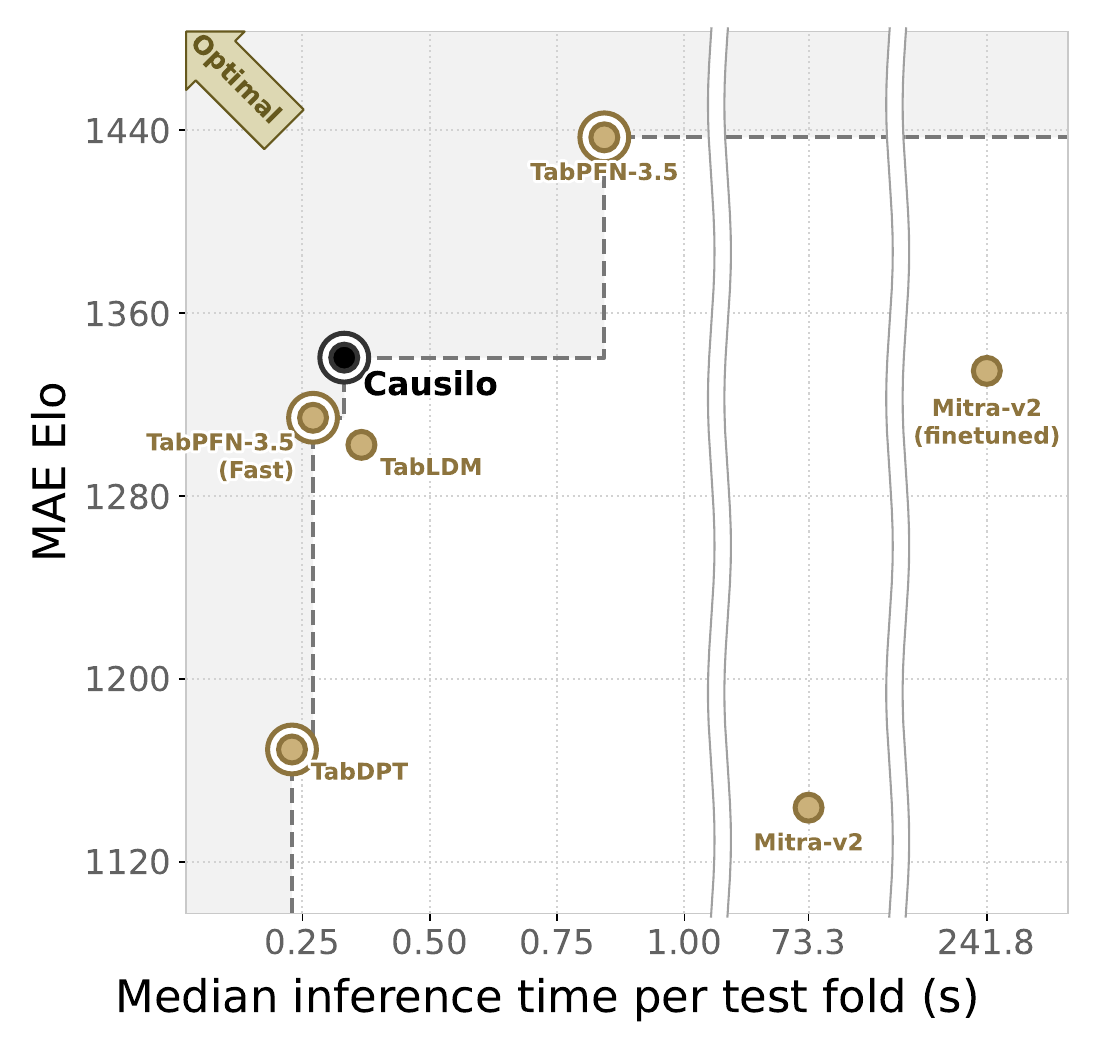}
    \caption{MAE.}
    \label{fig:scoringbench-mae}
  \end{subfigure}\hfill
  \begin{subfigure}[t]{0.485\textwidth}
    \centering
    \includegraphics[draft=false,width=\linewidth]{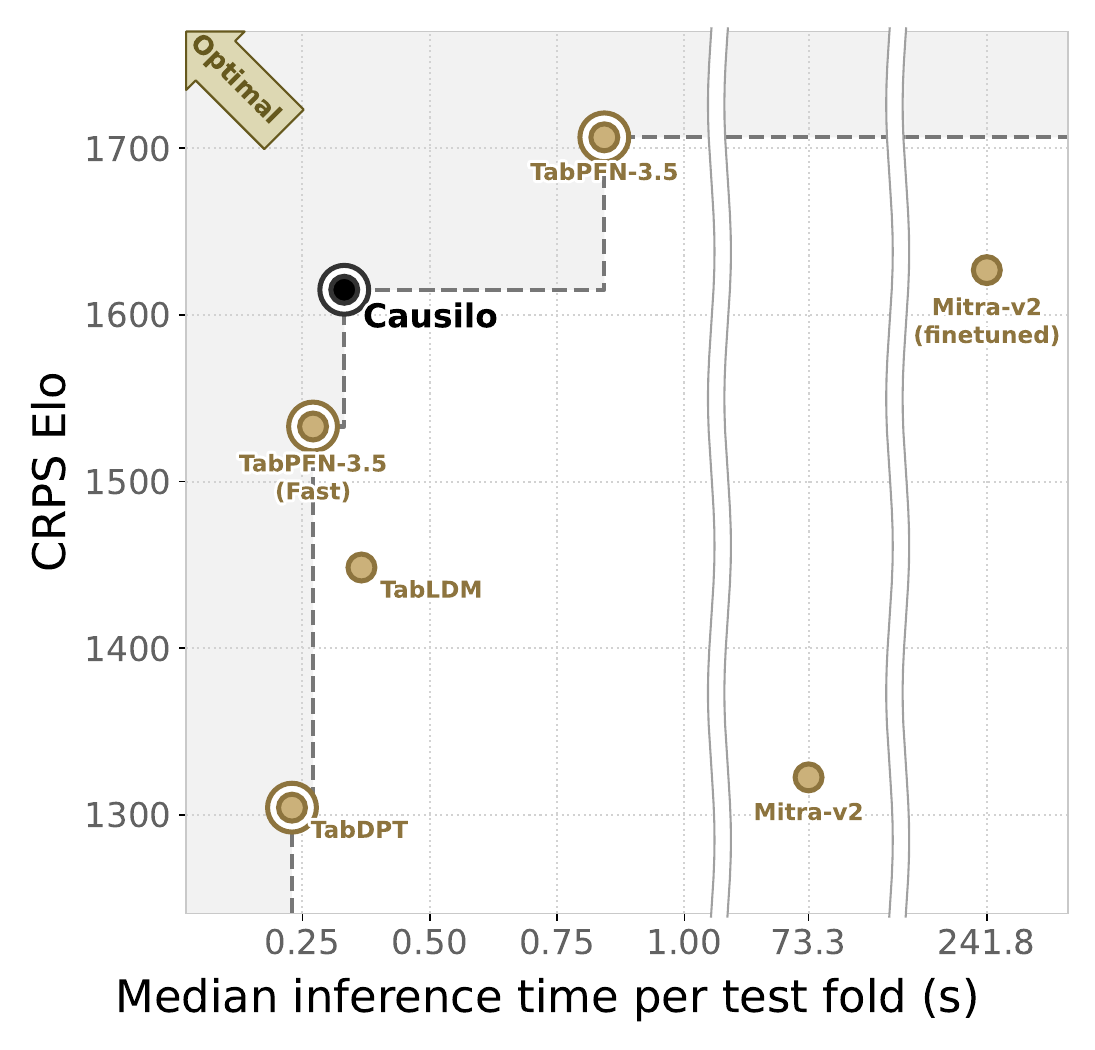}
    \caption{CRPS.}
    \label{fig:scoringbench-crps}
  \end{subfigure}
  \caption{ScoringBench performance--efficiency frontiers for point prediction (RMSE, $R^2$, and MAE) and distributional prediction (CRPS). Each panel plots metric-specific Elo against median inference time per test fold in seconds.  Dashed lines trace the empirical Pareto frontier; breaks in the time axis separate the much slower Mitra-v2 configurations. \method{} lies on the frontier in all four panels.}
  \label{fig:scoringbench}
\end{figure}

ScoringBench~\cite{landsgesell2026scoringbench} evaluates both point predictions and predictive distributions on a heterogeneous collection of real-world regression datasets. Figure~\ref{fig:scoringbench} compares metric-specific Elo ratings for RMSE, $R^2$, MAE, and CRPS against median inference time per test fold. Higher Elo indicates stronger predictive performance for every metric, while lower inference time is better. The broken time axis accommodates the substantially longer runtimes of the Mitra-v2 configurations.

\mypar{Point prediction.}
The three point-prediction metrics yield similar performance--efficiency frontiers. \method{} achieves higher Elo than TabPFN-3.5-Fast with a modest increase in inference time, while TabPFN-3.5 provides the highest Elo at a higher latency. \method{} also outperforms fine-tuned Mitra-v2 on all three point metrics at substantially lower inference cost. In the MAE panel, it additionally improves on TabLDM at lower latency, extending its advantage beyond squared-error metrics. It therefore occupies an intermediate point on the frontier between TabPFN-3.5-Fast and TabPFN-3.5, improving point-prediction quality without requiring the latency of the latter.

\mypar{Distributional prediction.}
CRPS evaluates the full predictive distribution by integrating the squared difference between its cumulative distribution function and the step function at the observed target. It rewards accurate distributions that are concentrated without being overconfident. Lower raw CRPS is better, corresponding to higher CRPS Elo in the figure.

\method{} remains on the CRPS frontier, outperforming TabPFN-3.5-Fast while requiring only modestly more inference time. TabPFN-3.5 again attains a higher Elo at a higher latency. Fine-tuned Mitra-v2 slightly exceeds \method{} in CRPS Elo, but requires a median 241.8 seconds per test fold, compared with approximately 0.3 seconds for \method{}. Across all four metrics, \method{} combines sub-second inference with competitive point and distributional prediction quality.

\section{License}
\label{sec:license}

The Causilo code is released under the Apache License 2.0.
The pretrained model weights are distributed separately under the Causilo License v1.0, which permits non-commercial research, testing, evaluation, experimentation, and modification.

Commercial or production use of the model, its derivatives, or its
outputs requires a separate license from Nums AI Inc.\ and any other
relevant rights holders. Providing the model or its derivatives
through hosted, managed, API, or SaaS services also requires a
license, whether the service is paid or free.

Scholarly publication and the sharing of research results, including outputs generated in compliance with the license, are permitted.
Independently authored papers and outputs need not adopt the model
license; any included model components remain subject to its
redistribution requirements. The full license texts govern all use
and distribution. Licensing inquiries may be directed to
\texttt{contact@nums.world}.
\section{Conclusion}

We introduced \method{}, a tabular foundation model that combines strong predictive performance with fast inference.
By introducing an additional refinement stage before row compression, \method{} preserves richer feature-level interactions while keeping  computation efficient.
Across TabArena, BeyondArena, and ScoringBench, \method{} establishes a strong position on the performance--efficiency frontier.

\FloatBarrier

\clearpage
\bibliographystyle{unsrt}
\bibliography{refs}

\clearpage
\appendix
\section{Contributors}
\label{sec:contributors}

Minyong Cho, Minho Jeong, Dooho Lee, Jinmo Lee,
and Jaemin Yoo.

The listing of contributors is in alphabetical order based on their last names.

\section{Acknowledgements}
\label{sec:acknowledgements}

We thank Prior Labs for their contributions to tabular foundation models and continued efforts to advance the field. We are grateful to the TabICL authors for openly sharing their code, models, and technical insights, which informed the development of \method{}. We also thank the authors and maintainers of TabArena, BeyondArena, and ScoringBench for creating and operating benchmarks that enable rigorous evaluation and meaningful comparisons across tabular learning methods.

\end{document}